\documentclass[sigconf]{acmart}

\AtBeginDocument{%
  \providecommand\BibTeX{{%
    \normalfont B\kern-0.5em{\scshape i\kern-0.25em b}\kern-0.8em\TeX}}}

\setcopyright{acmcopyright}
\copyrightyear{2020}
\acmYear{2020}
\setcopyright{acmcopyright}\acmConference[MM '20]{Proceedings of the 28th ACM International Conference on Multimedia}{October 12--16, 2020}{Seattle, WA, USA}
\acmBooktitle{Proceedings of the 28th ACM International Conference on Multimedia (MM '20), October 12--16, 2020, Seattle, WA, USA}
\acmPrice{15.00}
\acmDOI{10.1145/3394171.3413784}
\acmISBN{978-1-4503-7988-5/20/10}

\usepackage{multirow}

\newcommand{\myparagraph}[1]{\vspace{0.1em}\noindent\textbf{#1}}

\begin{document}
\fancyhead{}

\title{LGNN: A Context-aware Line Segment Detector}

\author{Quan Meng}
\email{mengquan@shanghaitech.edu.cn}
\affiliation{%
  \institution{ShanghaiTech University}
  \city{Shanghai}
  \country{China}
}

\author{Jiakai Zhang}
\email{zhangjk@shanghaitech.edu.cn}
\affiliation{%
  \institution{ShanghaiTech University}
  \city{Shanghai}
  \country{China}}

\author{Qiang Hu}
\email{huqiang@shanghaitech.edu.cn}
\affiliation{%
  \institution{ShanghaiTech University}
  \city{Shanghai}
  \country{China}
}

\author{Xuming He}
\email{hexm@shanghaitech.edu.cn}
\affiliation{%
 \institution{Shanghai Engineering Research Center of Intelligent Vision and Imaging, School of Information Science and Technology, ShanghaiTech University}
 \city{Shanghai}
 \country{China}}

\author{Jingyi Yu}
\email{yujingyi@shanghaitech.edu.cn}
\affiliation{%
  \institution{Shanghai Engineering Research Center of Intelligent Vision and Imaging, School of Information Science and Technology, ShanghaiTech University}
  \city{Shanghai}
  \country{China}}

\renewcommand{\shortauthors}{Trovato and Tobin, et al.}

\begin{abstract}
We present a novel real-time line segment detection scheme called Line Graph Neural Network (LGNN). Existing approaches require a computationally expensive verification or postprocessing step. Our LGNN employs a deep convolutional neural network (DCNN) for proposing line segment directly, with a graph neural network (GNN) module for reasoning their connectivities. Specifically, LGNN exploits a new quadruplet representation for each line segment where the GNN module takes the predicted candidates as vertexes and constructs a sparse graph to enforce structural context. Compared with the state-of-the-art, LGNN achieves near real-time performance without compromising accuracy. LGNN further enables time-sensitive 3D applications. When a 3D point cloud is accessible, we present a multi-modal line segment classification technique for extracting a 3D wireframe of the environment robustly and efficiently.

\end{abstract}



\maketitle

\section{Introduction}
\begin{figure}[t]
	\centering
	\includegraphics[width=0.9\linewidth]{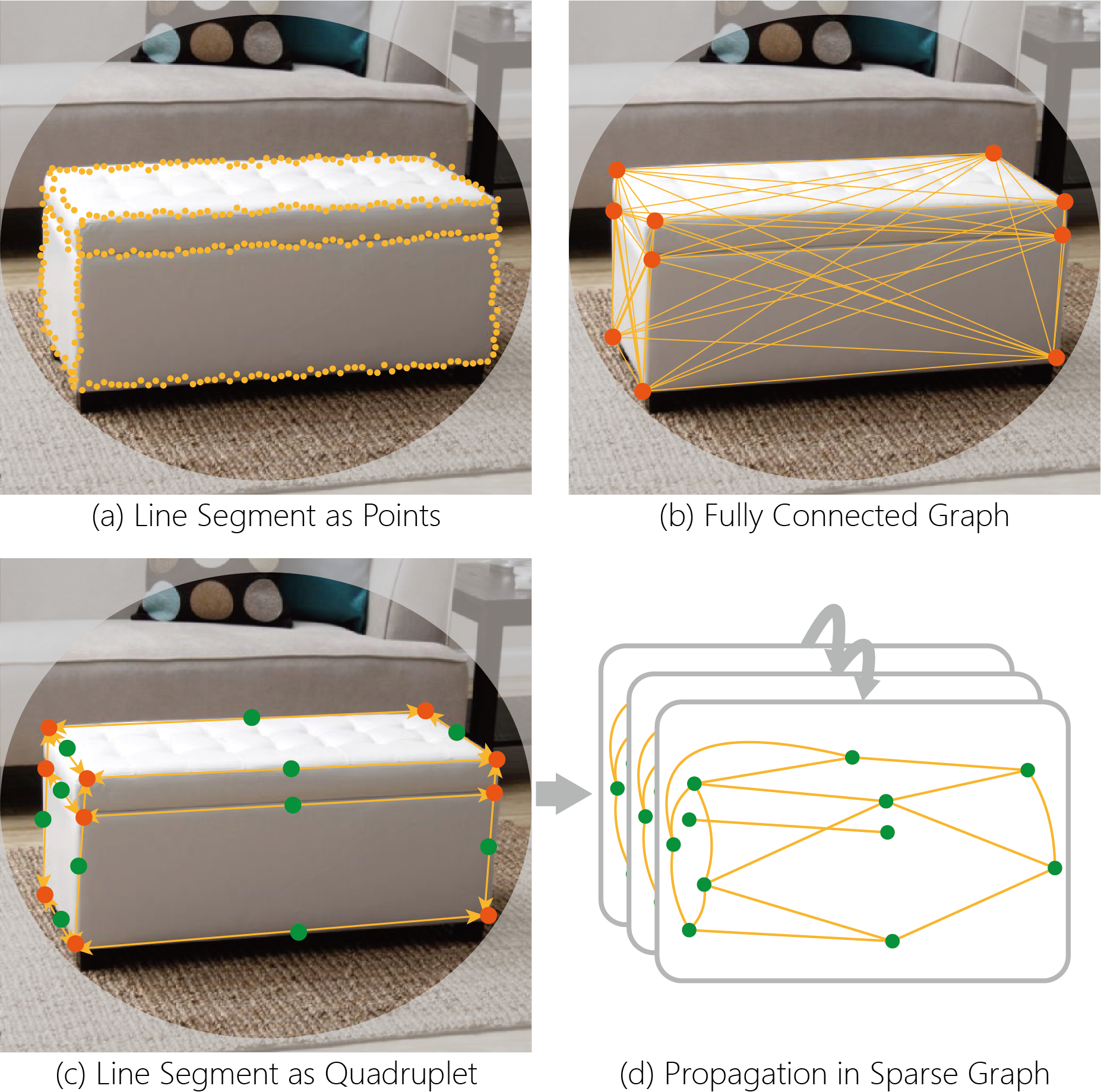}
	\caption{Different representations of 2D wireframe: (a) Grouping pixels to construct line segments~\cite{wireframe, afm}, which use sophisticated postprocessing and tend to produce short and overlapped line segments. (b) Modeling wireframe as a fully connected graph on line segment candidates~\cite{lcnn, ppgnet}, which are time- and memory-consuming and prone to ineffective inference due to the noisy proposals. (c \& d) Our proposed method of representing line segments as quadruplets: (start junction, end junction, line central point, line shift vector), which enables us to construct a sparse graph on line segment proposals and learn high-level semantic and geometric features during message-passing inference. }
	\Description{Different representations of wireframe}
\end{figure}

Line segments provide rich information about a scene: creases are indications of foldings of pliable surfaces, occlusion boundary edges encode shape information, while textures manifest the appearance of regions. More importantly, they provide a more precise, compact, and structural representation of a 3D scene. The detected line segments further benefit numerous computer vision tasks, ranging from stereo matching~\cite{yu2013line} and 3D reconstruction~\cite{hofer2017efficient, parodi19963d, zhang2014structure, denis2008efficient, faugeras1992depth} to image stitching~\cite{xiang2018image} and segmentation~\cite{arbelaez2010contour, de2007line}. Traditional techniques~\cite{lsd, mcmlsd, linelet, furukawa2003accurate, kamat1998complete, xu2015statistical} based on hand-crafted features are vulnerable to textureless regions, repetitive textures, illumination variations, occlusions, etc. More recent deep learning approaches~\cite{wireframe, afm, lcnn, ppgnet} attempt to explore semantic meanings of line segments to mitigate the problems. 

Existing learning-based algorithms tackle the line detection problem via a predict-then-verify strategy. Pioneering approaches~\cite{wireframe, afm} first adopt a deep convolution neural network (DCNN) to predict junctions as well as a line heat map or an attraction field map. They then apply sophisticated fusion algorithms for extracting line segments. Such approaches commonly produce crossing or fragmented line segments that are difficult to fix or even differentiate. More recent methods, including PPGNet~\cite{ppgnet} and L-CNN~\cite{lcnn}, first train a deep CNN to estimate a junction heatmap and then enumerate all junction pairs to verify their connectivities. The verification step greatly improves line detection quality but is time and memory consuming and scales poorly with the number of junctions in an image. For example, on a Tesla P40 GPU, verification over 512 junctions in PPGNet~\cite{ppgnet} requires about a second. 

For many real-life line detection applications, it is critical to balance between speed and performance. In this paper, we propose a real-time line detector --- Line Graph Neural Network (LGNN). LGNN can reliably handle a cluttered environment by exploiting a strong contextual structure between line segments. Specifically, LGNN employs two main modules: a DCNN module for generating line segment positions and features and a graph neural network for reasoning their connectivities. We propose a novel quadruplet representation - (start junction, end junction, line central point, line shift vector) - for each line segment, in place of the traditional junction-junction pairs. The DCNN sets out to predict a line central point heatmap along with a line shift vector map. We observe that, for cluttered scenes, the predicted line segments are less fragmented, where we can reliably map their endpoints to junctions. The GNN module then takes these line segment candidates as vertexes and construct a sparse graph to enforce structural constraints. 

Our LGNN significantly accelerates the detection speed without compromising accuracy. We show that LGNN achieves near real-time performance. On the wireframe dataset~\cite{wireframe}, LGNN performs at 15.8 frames per second (FPS) with $62.3\%$ structural AP (sAP) and a lightweight version achieves 34 FPS with $57.6\%$ sAP. LGNN hence enables time-sensitive 3D applications: when a 3D point cloud is accessible and we can map the predicted 2D line segments onto 3D to determine their types - creases, occlusion edges or texture edges. We therefore further present a multi-modal edge classification technique for extracting a 3D wireframe of the environment robustly and efficiently. 

\section{Related Works}
\myparagraph{2D Line Segment Detection.} Line segment detection has attracted a lot of research work. Classical approaches~\cite{lsd, mcmlsd, linelet, furukawa2003accurate, kamat1998complete, xu2015statistical} rely on low-level information, so are susceptible to external conditions. Recently, Wireframe~\cite{wireframe} first adopts two independent networks to predict line and junction heatmaps parallelly, then combine junctions and lines to produce line segments. AFM~\cite{afm} re-formulates it as a region coloring problem and leverage semantic segmentation networks to predict attraction field map, then group active pixels to construct line segments with region-growing algorithms similar to LSD~\cite{lsd}. Both of them need a sophisticated postprocessing method and tend to produce short line segments because they represent a line segment as a group of pixels. PPGNet~\cite{ppgnet} supplements the line segment dataset with outdoor scenes. L-CNN~\cite{lcnn} proposes line sampling to overcome the data unbalance, and a more reasonable metric(sAP) to evaluate the structural quality of wireframes. Both PPGNet and L-CNN represent line segments with endpoints and enumerate all junction pairs, so they scale poorly with the time complexity of $O(n^2)$. In this work, by representing line segments as quadruplets, our method not only directly get accurate line segments but also run the fastest.

\myparagraph{Objection Detection.} Object detection approaches contain two types of pipelines, namely, region proposal based and regression based approaches. The former approaches like R-CNN~\cite{girshick2014rich}, Fast R-CNN~\cite{girshick2015fast}, Faster R-CNN~\cite{ren2015faster}, R-FCN~\cite{dai2016r}, Mask R-CNN~\cite{he2017mask} and etc, generate region proposals at first and then classify each proposal into different object categories. The latter approaches like YOLO~\cite{redmon2016you}, SSD~\cite{liu2016ssd}, CornerNet~\cite{law2018cornernet}, CenterNet~\cite{zhou2019objects, duan2019centernet} and etc, remove the RoI extraction process and directly classify and regress the candidate anchor boxes. While the performance of region proposal based approaches remain in a higher place, they need more computation and processing time. Our line segment detection approach inherits merits of both approaches, in which we first predict line central points and junctions and directly regress other properties, followed by a light-weight GNN refinement.

\myparagraph{Graph Neural Network. }
Graph Neural Networks can effectively cope with non-Euclidean data, such as e-commerce~\cite{monti2017geometric, ying2018graph}, citation network~\cite{kipf2016semi, velivckovic2017graph}, molecules~\cite{defferrard2016convolutional, gilmer2017neural}, scene relationship~\cite{yang2018graph, yang2019sognet}, sketch recognition~\cite{xu2019multi} and etc. Recently, \citet{li2019deepgcns} build a very deep 56-layer Graph Convolutional Network(GCN) which significantly boosts performance in the task of point cloud semantic segmentation. Computer vision is one of the biggest application areas for graph neural networks. \citet{yang2018graph} propose a novel scene graph generation model called Graph R-CNN that reports state-of-the-art performance. \citet{xu2019multi} represent sketches as multiple sparsely connected graphs and designs the Multi-Graph Transformer that outperforms all RNN-based models. \citet{lu2019graph} first apply GNN in image segmentation and achieve about 1.34\% improvement on the VOC dataset compared to the FCN model. We are first to formulate wireframe as a sparse graph and enable message passing. This view allows us to encode larger context information so that we can decide which line segment is globally significant.

\begin{figure*}[th]
	\centering
	\includegraphics[width=1\linewidth]{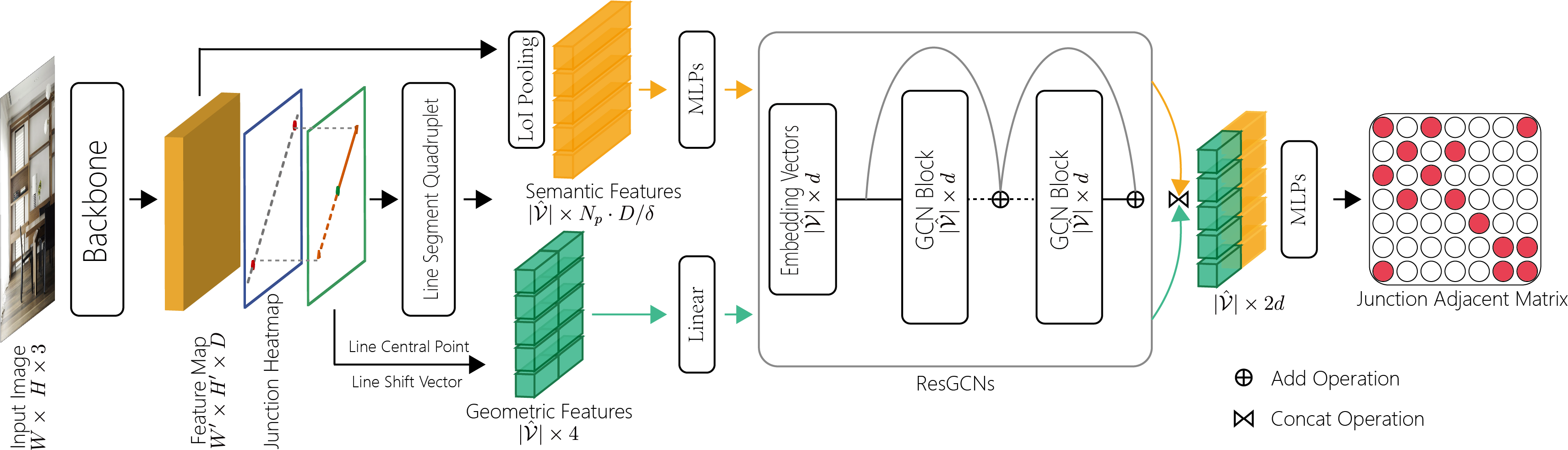}
	\caption{An overview of the proposed LGNN. Our model first uses a convolutional backbone network to produce a heatmap for line central point, a heatmap for junction, and a line shift vector map, which are assembled into a sparse graph of line segment quadruplets. We then introduce a deep ResGCNs to update the semantic and geometric features of line segments simultaneously. The resulting line representations are fed into an MLPs to score each line segment.}
	\label{fig:lgnn}
\end{figure*}

\myparagraph{3D Line Segment Detection.} Existing 3D line segment detection methods extract 3D line segments from an unordered point cloud. These methods mainly include three categories: point based~\cite{lin2017facet}, plane based~\cite{sampath2009segmentation} and image based~\cite{lin2015line, lu2019fast}. Given a point cloud, point based method identifies the boundary points and fits for the 3D line segment. This method usually mistakes much noise for boundary points. Plane based method detects planes and intersects every two adjacent planes to calculate line segments. This method may fail to find the terminals of the intersection line. Image based methods project point cloud into images and apply 2D line segment detector to extract line segments, finally re-project them to the point cloud. These methods only use hand-crafted features, so tend to produce short and crossed line segments, which are not good representations of the scene. In this work, we implement a structural 3D wireframe extraction algorithm based on LGNN. We also extract plane information to optimize the 3D line segment.

\section{Methods}
\subsection{Overview}
In this section, we introduce our line segment detection framework, which aims to delineate salient 2D line boundaries in an image. Our goal is to achieve high efficiency by exploiting rich properties of line segments, which enable us to effectively reduce the search space in localization and simplify the model structure. To this end, we develop a novel line segment detection method that consists of two main modules: a Multitask Learner Module and a Relation Reasoning Module. Given an input RGB image, the first module (Multitask Learner) is a multi-head deep convolutional network that extracts several key properties of line segments, such as their endpoints and orientations. We then generate a set of line segment and junction proposals, and construct a \textit{sparse graph} on line candidates with junctions as graph edges. The second module (Relation Reasoning) is a graph convolutional network that augments the line features with context cues, which are subsequently used for final line segment prediction. An overview of our framework is shown in Fig.~\ref{fig:lgnn}. 

The rest of this section first introduces our line representation in Section~\ref{sec:rep}, followed by two model components with inference process in Section~\ref{sec:mlnet} and Section~\ref{sec:rrnet}. Finally, we describe our training strategy and multi-task loss function in Section~\ref{sec:train}.      


\subsection{Line Segment Representation} \label{sec:rep}

Given an RGB image $I \in \mathbb{R}^{W \times H \times 3}$, we first denote the set of line endpoints or junctions as $\mathcal{J}$ and the set of line central/middle point as $\mathcal{C}$. In order to generate line segment proposals efficiently, we represent a line segment as a quadruplet $v = (j^1, j^2, c, \mathbf{s})$, in which $j^1$ and $j^2\in\mathcal{J}$ are the two endpoints of the line segment $v$, $c\in\mathcal{C}$ is its central point, and $\mathbf{s}\in\mathbb{R}^2$ is a 2D shift vector indicating the line direction and segment length. 

Concretely, we denote the 2D coordinates of the endpoints $j^1, j^2$ and the central point $c$ as $\mathbf{p}_{j^1}$, $\mathbf{p}_{j^2}$, $\mathbf{p}_{c}\in\mathbb{R}^2$, respectively. The 2D shift vector $\mathbf{s}$ of the line central point $c$ satisfies the following relations:
\begin{equation}
\setlength{\abovedisplayskip}{3pt}
\setlength{\belowdisplayskip}{3pt}
\label{eq:line}
({\mathbf{p}}_{{j}^1}, {\mathbf{p}}_{{j}^2}) = ({\mathbf{p}}_{{c}}-{\mathbf{s}}, {\mathbf{p}}_{{c}}+{\mathbf{s}})
\end{equation}
To avoid ambiguity of directions, we stipulate that $\mathbf{s}$  is always pointing to the endpoint closer to the right side of the image.

To capture relations between line segments, we further represent the entire set of line segments in an image as a  
graph $\mathcal{G} = (\mathcal{V}, \mathbf{A})$, where $\mathcal{V}$ stands for the set of the unordered line segments, $\mathbf{A}$ is the adjacent matrix representing the connectivity between these line segments. For two line segments, $v_k$ and $v_l$, if they have a common endpoint, i.e., they are connected, then both $\mathbf{A}_{kl}$ and $\mathbf{A}_{lk}$ equal one; and otherwise, they are zero. Our goal is to develop a deep network to predict the graph $\mathcal{G}$ from the input image $I$, which will be described in detail below. 

\subsection{Mutlitask Learner Module}\label{sec:mlnet}
Our first module, the Multitask Learner, takes the image $I$ as input and generates an initial estimation of line segment properties including their endpoints, central points and shift vectors. To achieve this, we develop a multi-head convolutional network with two components: a backbone ConvNet for feature extraction and a prediction module that outputs three properties of the line segments.  

In this work, we adopt a stacked hourglass network~\cite{newell2016stacked} as our backbone. Using multiple bottom-up and top-down inference across scales, this backbone network produces a rich set of feature maps $\mathbf{F} \in \mathbb{R}^{W' \times H' \times D}$, where $W'$, $H'$ and $D$ are the width, height and the number of channels of the feature maps. 

The prediction module consists of three parallel convolutional heads, generating dense output maps for endpoints, central points and shift vectors, respectively. Each of the endpoint and central point heads produces an output map that indicates the confidence scores of each location belonging to the junction set $\mathcal{J}$ or the line central point set $\mathcal{C}$. We adopt a heatmap representation for those keypoints as in the human pose estimation. Besides, due to the discretization effect caused by the output stride, we also produce two offset maps to predict offsets for junctions and line central keypoints. 
The shift vector head simultaneously regresses the direction and (half) length of the line segments for each location. 

We apply non-maximum suppression to remove duplicated keypoints and extract the local maximum in the line central point heatmap as candidates. To generate consistent line segment and endpoint/junction proposals, we introduce a simple nearest neighbor-based alignment procedure to match the predicted endpoints to the pairs of central point and shift vector and remove noisy center point candidates as they are less reliable.

Specifically, given a central point candidate $\hat{c}$ and its shift vector, $\hat{\mathbf{s}}$, we first generate a 2D line segment proposal $\hat{v}$ with endpoints $\hat{j}^1$ and $\hat{j}^2$ by shifting the line central point $\hat{c}$ in two opposite directions with the vector $\hat{\mathbf{s}}$ as in Eq.~\ref{eq:line}. We then match the endpoints of $\hat{v}$ to the predicted endpoint candidates. In particular, we find the closest endpoints to each generated line segment. If the total distance is below a threshold $\theta$, we will replace the computed endpoints $\hat{j}^1$ and $\hat{j}^2$ by the matched endpoint candidates. If the distance exceeds $\theta$, we will remove the line segment candidate. The isolated endpoint candidates will also be filtered out after matching.    
Finally, we build a candidate graph $\mathcal{\hat{G}}=(\mathcal{\hat{V}}, \hat{\mathbf{A}})$ using the remaining line segment and endpoint candidates, in which the vertex set $\hat{\mathcal{V}}$ comprises line segments and the adjacent matrix $\hat{\mathbf{A}}$ encodes the connectivity between lines due to shared endpoints.   



\subsection{Relation Reasoning Module}\label{sec:rrnet}

Given the candidate graph $\mathcal{\hat{G}}$, we now introduce our Relation Reasoning Module, which is a graph neural network defined on the top of $\mathcal{\hat{G}}$. Each vertex in the graph neural network is associated with a line segment proposal and is connected to other line proposals based on the adjacent matrix $\hat{\mathbf{A}}$. The graph neural network takes line segment features as input and conducts global reasoning through message passing, resulting in a context-aware representation for each line proposal. We then predict a binary label for each graph vertex to indicate whether it is a foreground line segment.      

Concretely, we first extract two sets of line segment features to encode their semantic and geometric property. For semantic features, we adopt the LoI pooling~\cite{lcnn}, which max-pools and concatenates convolutional features from a set of uniformly sampled points on the line segments. Let the number of sampled points be $N_p$ and the pooling stride be $\delta$, we denote the semantic feature of vertex $v\in\mathcal{\hat{V}}$ as $\hat{\mathbf{x}}_v^s \in \mathbb{R}^{{N_p\cdot D}/{\delta}}$. For geometric features, we concatenate the line central point's coordinate ${\mathbf{p}}_{\hat{c}}$ and the shift vector $\hat{\mathbf{s}}$, which is denoted as $\hat{\mathbf{x}}_v^p$.



Given the input features, we initialize the vertex representations of graph neural network by computing an embedding of line segment features:
\begin{equation}
\mathbf{e}_v^{(0)} = \Phi(\hat{\mathbf{x}}_v^s), \quad \mathbf{g}_{v}^{(0)} = \Psi(\hat{\mathbf{x}}_v^p),\; v\in\mathcal{\hat{V}} 
\end{equation}
where $\mathbf{e}_{v}^{(0)}$ and $\mathbf{g}_{v}^{(0)}\in \mathbb{R}^d$ are the embedded representations of the semantic and geometric features of the vertex $v$. Here $\Phi$ is a two layer perceptron, and $\Psi$ is simply a linear projection.

We update the semantic and geometric representations in parallel by running two separate message passing procedures in the graph in order to capture both semantic and geometric context. To achieve this, we first compile features from all the neighborhoods of vertices in our graph convolutional network and then perform a non-linear transformation on the aggregated features to update the representations of the vertices. To capture long-range context, we stack multiple layers of such graph convolutions.  

Specifically, we adopt the residual GCN blocks~\cite{li2019deepgcns} for message computation and vertex feature update within each layer. Let $\mathbf{E}=[\mathbf{e}_{1},\cdots,\mathbf{e}_{|\mathcal{\hat{V}}|}]$ and $\mathbf{G}=[\mathbf{g}_{1},\cdots,\mathbf{g}_{|\mathcal{\hat{V}}|}]$ be the collections of semantic and geometric representations of the vertices, we update their embedding in the $l+1$ layer as follows,
\begin{align}
\setlength{\abovedisplayskip}{3pt}
\setlength{\belowdisplayskip}{3pt}
\mathbf{E}^{(l+1)} &= \phi\left(\widetilde{\mathbf{D}}^{-\frac{1}{2}} \widetilde{\mathbf{A}} \widetilde{\mathbf{D}}^{-\frac{1}{2}} \mathbf{E}^{(l)} \mathbf{W}_s^{(l)}\right) + \mathbf{E}^{(l)}\\
\mathbf{G}^{(l+1)} &= \phi\left(\widetilde{\mathbf{D}}^{-\frac{1}{2}} \widetilde{\mathbf{A}} \widetilde{\mathbf{D}}^{-\frac{1}{2}} \mathbf{G}^{(l)} \mathbf{W}_g^{(l)}\right) + \mathbf{G}^{(l)}
\end{align}
where $\widetilde{\mathbf{A}} = \hat{\mathbf{A}} + \mathbf{I}$, $\widetilde{\mathbf{D}}$ is a diagonal matrix with $\widetilde{\mathbf{D}}_{i i}=\sum_{j} \widetilde{\mathbf{A}}_{i j}$.  $\mathbf{W}_s^{(l)}$ and $\mathbf{W}_g^{(l)}$ are the weight parameters in the $l$-th layer. $\phi$ is an activation function, and we simply choose ReLU funcion.

We stack $n$ residual GCN blocks to update the semantic and geometric features of vertices, which are concatenated to generate the final line segment representation: 
$\mathbf{H} = [\mathbf{E}^{(n)}, \mathbf{G}^{(n)}]\in \mathbb{R}^{|\mathcal{\hat{V}}| \times 2d}$.
Finally, we adopt a multi-layer perceptron to classify line segments into foreground or background based on the updated line segment features $\mathbf{H}$. During inference, we also attach a sigmoid function to generate the final score for each line segment.

\begin{figure*}[th]
	\centering
	\includegraphics[width=1.0\linewidth]{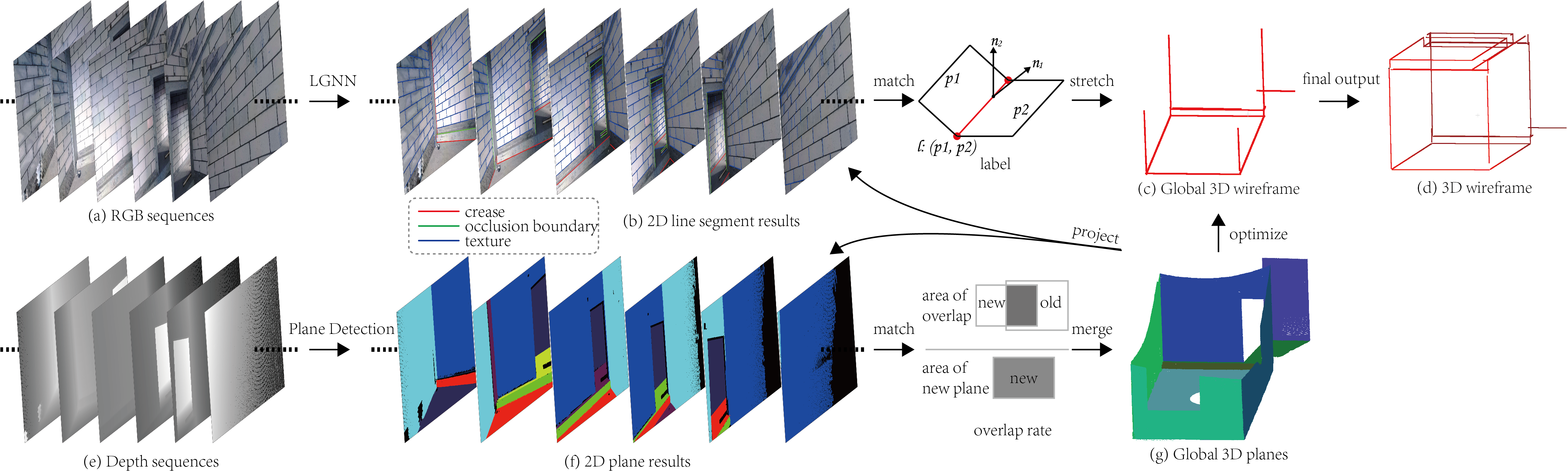}
	\caption{Flowchart of the 3D line segment detection. Given a sequence of RGBD images, we first detect 2D line segments and surface planes from RGB and depth, respectively. These results are subsequently fused into a 3D wireframe based on 2D-3D and line-plane consistency.}
	\label{fig:application_flow}
\end{figure*}

\subsection{Model Training}\label{sec:train}

To train our model, we develop a multi-task loss to supervise the learning of two model modules jointly. The loss function consists of two parts, one for the Multitask Learner Module and one for the entire network:
\begin{equation}
L = L_{ML} + L_{RR}   
\end{equation}
where $L_{ML}$ denotes the loss terms for the first module and the $L_{RR}$ is the loss terms imposed on the output of the second module.  

The loss term $L_{ML}$ includes the loss terms for the four outputs of the Multitask Learner Module as follows:
\begin{equation}
L_{ML} = \lambda_{j}L_j + \lambda_{c}L_c + \lambda_{o}L_o +  \lambda_{s}L_s   
\end{equation}
where the loss item $L_{j}$ is for junction keypoints, $L_c$ is for line central keypoints, $L_o$ is for offset of junction keypoints and line central keypoints, $L_s$ is for line shift vector, and $\lambda_{j, c, o, s}$ are the weights of the corresponding loss item.   

Specifically, we use the binary cross entropy loss $L_{j}$ for junction/endpoint prediction, and 
\begin{equation}
L_{j} = -\frac{1}{N_p}\sum_j(y_j \log (\hat{y}_j)+(1-y_j) \log (1-\hat{y}_j))
\end{equation}
where $y_j$ is the binary junction indicator and $\hat{y}_j$ is the predicted junction probability. $N_p$ is the number of pixels in the output heatmap.

For the line central point prediction, 
we use the focal loss~\cite{lin2017focal} for line central point estimation due to unbalanced positive/negative distribution:
\begin{equation}
L_{c}= -\frac{1}{N_p}\sum_c\left\{\begin{array}{cc}
\left(1-\hat{y}_c\right)^{\alpha} \log \left(\hat{y}_c\right) & \text { if } y_c=1 \\
\left(1-y_c\right)^{\beta}{\hat{y}_c}^{\alpha} \log \left(1-\hat{y}_c\right) & \text { otherwise }
\end{array}\right.
\end{equation}
where $y_c$ is ground truth scores of line central points and $\hat{y}_c$ is predicted line central point probability. $\alpha$ and $\beta$ are hyper-parameters. We use $\alpha = 2$ and $\beta = 4$ as in \cite{law2018cornernet}. 

We employ l1 loss for the junction and line central point offset regression loss $L_{o}$, and the shift vector prediction loss $L_s$. 

The loss term $L_{RR}$ is defined for the final outputs from the MLPs of the Relation Reasoning Module. Here we use the binary cross-entropy loss for line segments classification.

We train our network in a joint manner by computing an approximate gradient over the entire network, in which the gradient calculation treats the proposal generation is fixed in each iteration. This ``end-to-end" training strategy works well in practice for our model and is simple to implement. 

\section{3D Wireframe Extraction}
A unique advantage of our LGNN-based line segment detector is its speed. Compared with the most accurate technique~\cite{lcnn}, LGNN slightly sacrifices the performance but nearly doubles the speed. The near real-time performance can benefit a number of applications. 

When combined with 3D scanning, LGNN provides a viable solution for space measurement. Conceptually 3D scanning techniques such as LiDAR or time-of-flight can already produce 3D point cloud data amenable for analysis. In reality, the point cloud is generally of a low resolution and contains strong noise. We observe that walls, in particular, intersection between different walls, form line segments and as long as we can detect them, we can use the results for measuring the dimensionality of the space. 

However, there are multiple challenges. Our line detector, same as any existing techniques, detects both geometric and texture edges. For the detected lines to be useful, it is essential to distinguish these lines: the key to space measurement is creases that correspond to junctions of walls. We employ a real-time line type classification algorithm on top of LGNN. Fig.~\ref{fig:application_flow} shows our processing pipeline. We assume a moving 3D scanner whose position is calibrated in real-time using Visual SLAM techniques such as ORB-SLAM~\cite{mur2015orb}. The point clouds are fused on the fly. Given each line segment provided by LGNN, we check if it corresponds to crease. 

Specifically, we observe that it corresponds to crease when the two sides of the line segment correspond to two different planes (i.e., have very different normal directions). For texture lines, the two sides would correspond to the same plane. However, for occlusion boundaries, the two sides can also correspond to different planes as creases. Nonetheless, we can easily identify occlusion boundaries since the points on the two sides will have large depth disparity. Further, regular rooms have walls perpendicular to each other and therefore we can check if the normals are orthogonal to further improve accuracy on crease detection. 

Two key steps in the approach above are 1) to group 3D points in terms of planes that they belong to and 2) to stretch line segments as the camera moves, to avoid fragmentation. Fig.~\ref{fig:application_flow} shows the complete pipeline of our technique: using a sequence of RGBD frames as input, we use LGNN to extract, for every single frame, line segments. We maintain a global 3D plane set and a global 3D line set. To maintain the 3D plane set, we implement a point-plane merging technique that adds newly detected 3D points to the set. To maintain the 3D line set, we use the plane set to classify the 2D lines detected by LGNN and refine the set on the fine. Both steps can be implemented in real time. 



To elaborate, we apply LGNN on an RGB frame and the fast plane detection method~\cite{feng2014fast} on the depth channel. The difficulty is that once we move the camera, we need to merge newly detected points with the existing ones. Further, the RGBD sensor is generally of a low resolution and the new points will slightly deviate from the actual plane. To handle that, we record the total number of already scanned 3D points $N_i$ and label every point with a unique plane id $i$. When new 3D points become accessible, we add a plane $\Pi_{new}$ into global plane set with a point count $N_{new}$ (i.e., how many points in $\Pi_{new}$) and normal $n_{new}$. We set out to determine if $\Pi_{new}$ should be merged with the existing set or should be added as a new one. 

To do so, we project the global planes into the current RGB image via standard graphics rasterization. We then calculate its overlap count $N_{i}$ of projected global plane $\Pi_{i}$ with respect to the points in $\Pi_{new}$. We define overlap ratio $r_i = \frac{N_i}{N_{new}}$. Finally, we merge $\Pi_{new}$ with $\Pi_{max}$ that has the maximum overlap ratio with $\Pi_{new}$. If $r_{max}$ exceeds a threshold and the normals of $n_{new}$ and $n_{max}$ are close enough, we will merge them into one plane. Otherwise, we will create a new plane and add it to the global plane set. 

\begin{figure*}[ht]
	\centering
	\includegraphics[width=0.97\linewidth]{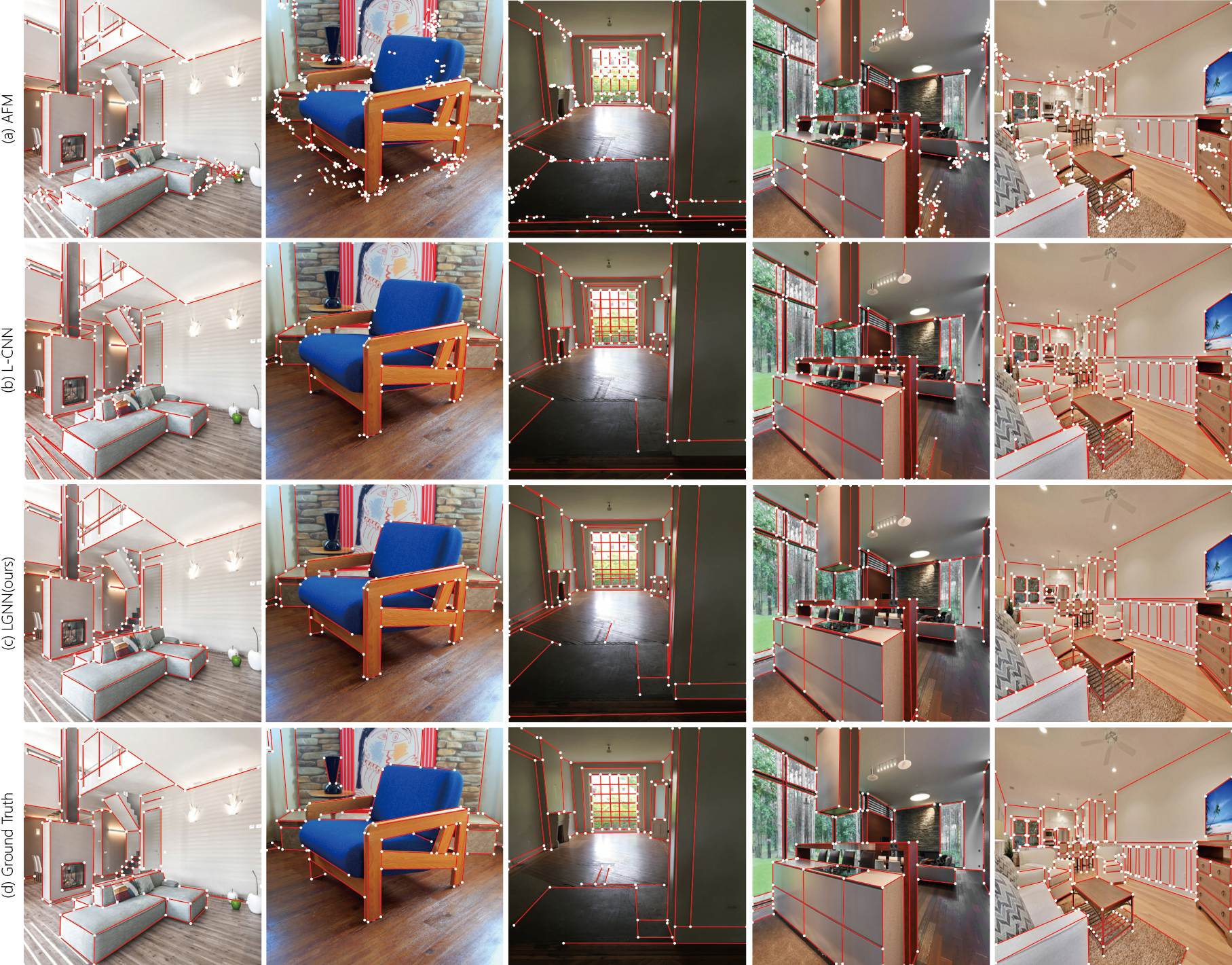}
	\caption{Qualitative results on 2D line segment detection. First row: AFM~\cite{afm}; Second row: L-CNN~\cite{lcnn}; Third row: LGNN(ours); Fourth row: Ground truth. Our method achieves competitive results with real-time efficiency.}
	\label{fig:results}
\end{figure*}

\begin{figure*}[ht]
	\centering
	\includegraphics[width=0.97\linewidth]{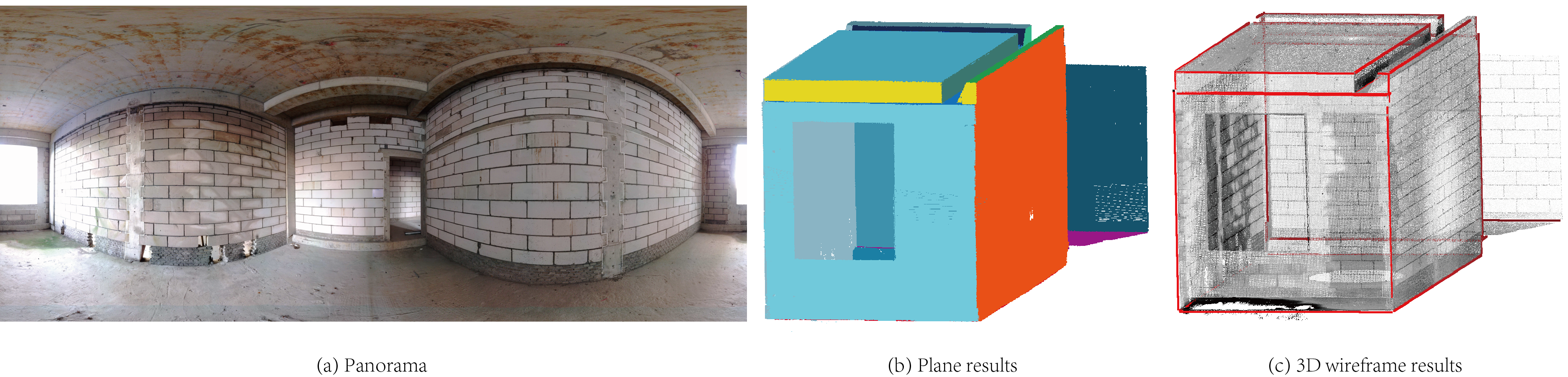}
	\caption{Given a sequence of RGBD images and camera poses as inputs, LGNN enables a 3D wireframe parsing algorithm to detect structural line segments in real-time, and to fuse with estimated planar surfaces for online 3D scene reconstruction.
	}
	\label{fig:teaser}
\end{figure*}

For line segment classification, we simply build a histogram in terms of the plane id of the points that lie within a range on both sides. If the histogram contains a very high peak that corresponds to a single id, the line is then deemed as a texture edge. If the histogram has two similar peaks of different ids and the depths of the points are similar, the line segment is then deemed as crease. Otherwise, it is deemed as an occlusion boundary. We further label the creases with the indexes of its two adjacent planes so that we can match these line segments in following frames. 

The classification procedure avoids fragmentation of line segments that correspond to creases: we can merge creases that correspond to the same pair of junction planes (i.e., the same plane ids). In fact, we can obtain the final line segment by simply computing the intersection line of the two planes. Fig.~\ref{fig:application_flow}(b) shows several typical examples of the line segment classification results. Finally, we can use the merged creases to obtain a wireframe model of the 3D environment and then measure the space, as shown in Fig.~\ref{fig:application_flow}(d). 




\section{Experiments}
In this section, we introduce details of our implementation and evaluate the proposed line segment detector with existing state-of-the-art line segment detectors. Then, we visualize the results of our 3D wireframe detection system.

\subsection{Implementation Details}
We stack two hourglass networks as our backbone, for each input image, we first resize it to the size of $(512, 512, 3)$ and output a feature map with the size of $(128, 128, 256)$. Then, we feed the intermediate feature map into five network heads and produce junction heatmap, line central point heatmap, junction offset map, line central offset map, and line shift vector map. During the training phase, the number of proposals of junctions is two times the number of ground truth junctions and with the maximum value of 300. During the evaluating period, we set a threshold of $0.008$ to choose the most likely junctions. To get line segment quadruplets, we set the threshold of $\theta$ of $15$ to get matched line segments. For the LoIPooling, we first uniformly choose $N_p = 32$ middle point features along each line segments, then apply a max-pooling with the stride of $4$ to get the flattened semantic line feature of size $2048$. For graph neural network, we first embed the semantic feature and geometric feature to the same size of $256$, then stack several residual GCN blocks to update the feature. Finally, we use a two-layer MLPs with hidden layers of the size of $32$ to classify each line segment.

We train LGNN from scratch using ADAM optimizer~\cite{kingma2014adam}, with an initial learning rate of $1.0e-3$ and weight decay of $1.0e-4$ on a single GPU P6000. We set batch size to $10$ for the fastest training speed. To dynamically adjust the learning rate based on validation measurements, we adopt the ReduceLROnPlateau\footnote{For eg. \url{https://pytorch.org/docs/stable/optim.html?highlight=reducelronplateau\#torch.optim.lr\_scheduler.ReduceLROnPlateau}} scheduler with the patience of zero epoch and factor of $0.5$. We augment the wireframe dataset with standard strategies including flipping vertically, horizontally, and centrally for images and annotations to overcome overfit. 

\subsection{Performance Evaluations}
We evaluate our line segment detection performance on the wireframe dataset~\cite{wireframe}, which contains $5,000$ training images, $462$ testing images. For faster training, we preprocess the wireframe dataset to generate ground-truth keypoint, offset and shift vector maps. Specially, we generate ground-truth line central point maps by using the 1D Gaussian kernel along each line segment. This step is crucial for our network to converge well.

\paragraph{LGNN vs. State-of-the-Art.}
We have compared our LGNN with the state-of-the-art line segment detection algorithms: AFM~\cite{afm} and L-CNN~\cite{lcnn}, with the same wireframe dataset~\cite{wireframe}, the same training and testing split, and the same hardware environments. 

We experiment with the sAP metric which is proposed by L-CNN~\cite{lcnn} to evaluate the performance of these methods. The sAP metric for line segment detection properly penalizes for the overlapped and incorrectly connected line segments, so is a more reasonable metric for evaluating the structural quality of wireframes compared with the heat map-based metric --- $AP^H$, which treats each pixel independently. As reported in Table~\ref{tab:compare_sAP}, our method achieves comparable results while runs the fast.

\begin{table}
  \begin{tabular}{ccc}
    \toprule
   Methods    & sAP          & FPS   \\
    \midrule
    Wireframe~\cite{wireframe} &     6.0      &  3.9      \\
    AFM~\cite{afm}             &    27.5      &  12.0     \\
    L-CNN~\cite{lcnn}            &    63.0      &  9.5      \\
    Ours-lite                   &    57.6         &  34.0      \\
    Ours                        &    62.3      &  15.8     \\
  \bottomrule
 \end{tabular}
	\caption{Performance comparison of line segment detection approaches on the wireframe dataset~\cite{wireframe}. We adopt sAP~\cite{lcnn} as our evaluation metric and report the average FPS on the test set of wireframe~\cite{wireframe}.}
	\label{tab:compare_sAP}
\end{table}

We visualize results of the proposed LGNN and other methods in Fig.~\ref{fig:results}. We can see that our approach is capable of extracting complete and cleaner line segments compared with AFM~\cite{afm} and L-CNN~\cite{lcnn}. Treating each line segment as a quadruplet, we get an accurate description of each line segment. Conversely, AFM generates line segment by greedily grouping pixels, this way usually fails to guarantee a complete and accurate line segment. Compared with L-CNN~\cite{lcnn}, although it gets the best sAP performance, it produces more overlapped or close co-linear line segments. Our method suppresses most of these line segments so that our results look cleaner.

\paragraph{Ablation Studies.}
In this section, we run several ablation experiments to study the Relation Reasoning Module in our proposed method:

(i) We experiment with several types of line segment features: only semantic feature; only geometric feature; both semantic and geometric feature; semantic feature, geometric feature, and keypoint scores. We can see that effectively combining semantic and geometric feature ensures the best line segment prediction performance. We also attempt to add the keypoint scores as an additional feature, which however can not further improve the performance. We conjecture that this might be caused by the redundancy and noisy nature of the keypoint scores.

\begin{table}[t]\centering
	\begin{tabular}{ccccc}
		\toprule
		\multirow{2}[3]{*}{semantic} & 
		\multicolumn{2}{c}{geometric} &
		\multirow{2}[3]{*}{scores} &
		\multirow{2}[3]{*}{sAP} \\
		\cmidrule(lr){2-3}
		& coord & shift &  &  \\
		\midrule
		 \checkmark  &   &   &   & 61.4 \\
		 \checkmark & \checkmark &   &   & 61.7 \\
		 \checkmark & \checkmark & \checkmark &   & 62.3 \\
		 \checkmark & \checkmark & \checkmark & \checkmark & 61.4 \\
		\bottomrule
	\end{tabular}
	\caption{Ablation study on multiple features in the Relaton Reasoning Module. 'coord' represents the line central point position, 'shift' represents the line shift vector, 'scores' represents the 3d concatenated keypoint scores of the line quadruplet.}
	\label{gnn feat}
	\Description{Ablation study for multi-feature for Relaton Reasoning Module.}
\end{table}

\begin{figure}[ht]
	\centering
	\includegraphics[width=0.7\linewidth]{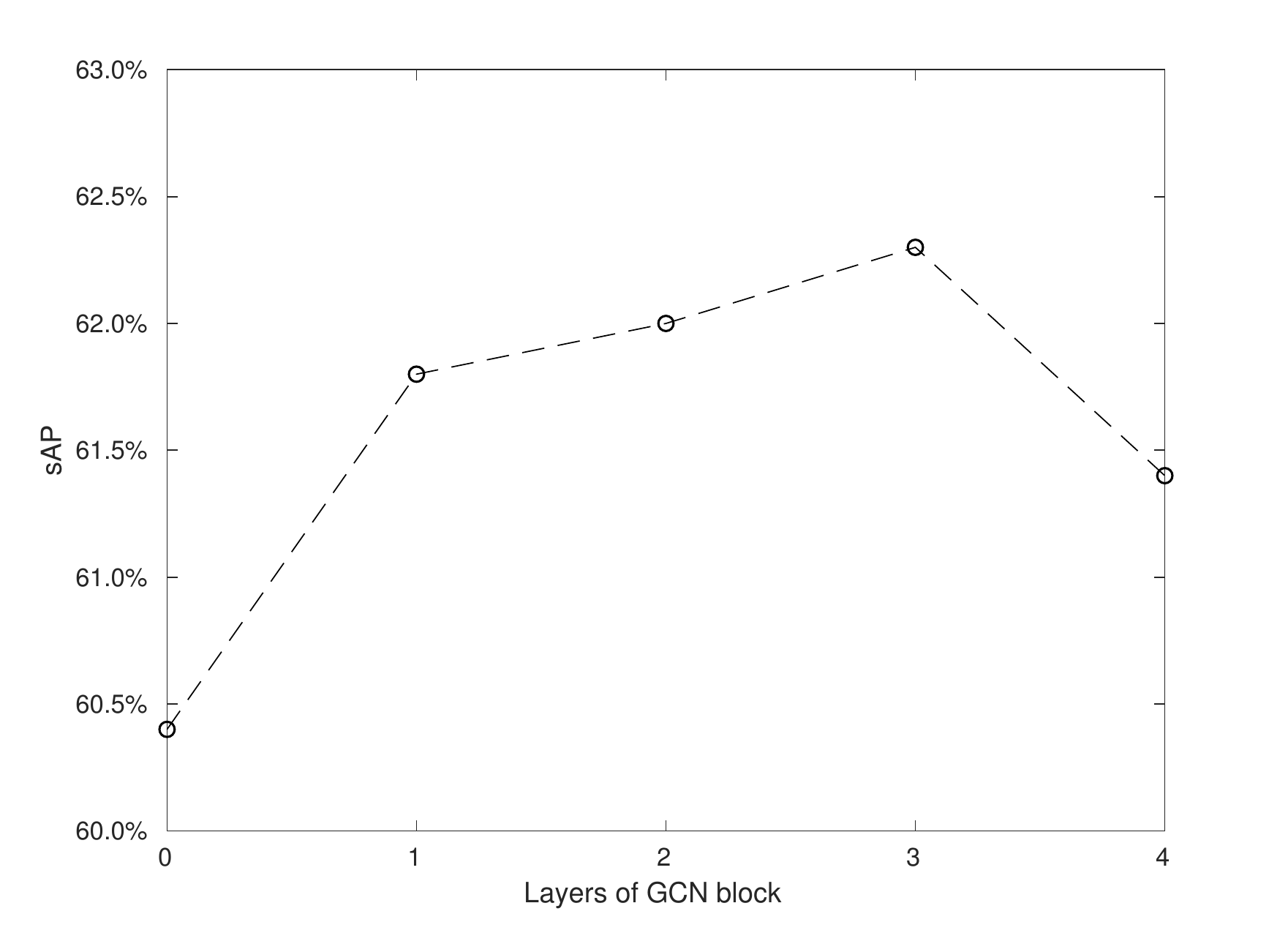}
	\caption{Model performance using different layers of ResGCNs in the Relation Reasoning Module. We show the sAP on the wireframe~\cite{wireframe} test set for 0, 1, 2, 3, and 4 layer ResGCNs.}
	\label{gnn layers}
	\Description{Training different layers of ResGCNs.}
\end{figure}

(ii) We also evaluate our Relation Reasoning Module with different layers of ResGCNs. In Fig.~\ref{gnn layers}, we achieve the best performance with three layers ResGCNs. The performance of shallower ResGCNs shows a steady decrease. When the number of layers drops to zero, i.e., does not pass message, there is a sharp performance gap. Compared with MLPs, ResGCNs obtains an absolute gain of $1.9\%$ in terms of sAP. The deeper ResGCNs also has a lower sAP partially due to the difficulty in model learning.  

For a fair comparison among CNN, MLP, and GCN, we add CNN layers to the CNN backbone or substitute GCN with MLP in our proposed Relation Reasoning Module so as to keep the same layers and the same feature dimensions. We have experimented that GCN obtains an absolute gain of $1.9\%$ in terms of sAP, while MLP and CNN only obtain $1.1\%$ and $0.0\%$, respectively. We conclude that line segment detection benefits greatly from ResGCNs which can more effectively aggregate and cope with context information.

\paragraph{3D Wireframe Extraction.}
\begin{figure}[ht]
    \centering
    \includegraphics[width=1.0\linewidth]{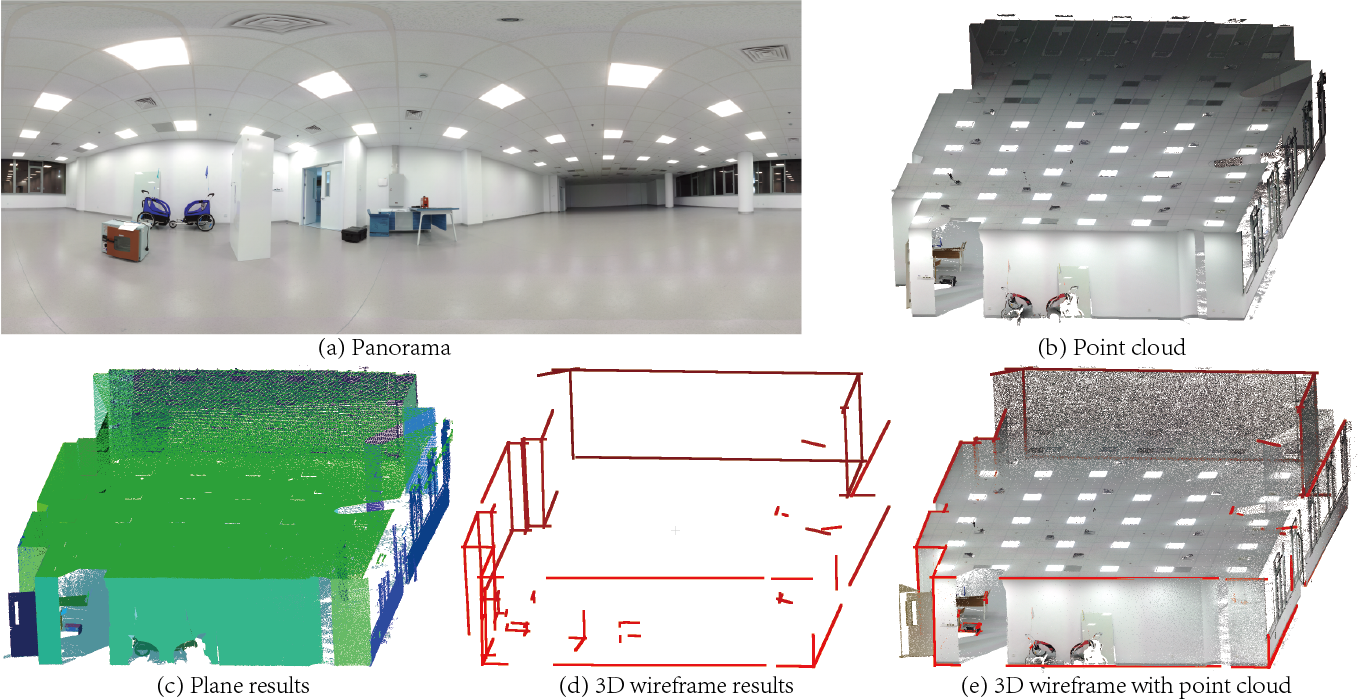}
    \caption{Wireframe parsing results on a large robotic lab. (a) Illustration of the overall scene as a panorama. (b \& c) Point cloud and results of plane detection. (d \& e) 3D wireframe result and its overlay with the point cloud.}
    \label{fig:applicationresults2}
\end{figure}

We have further tested our 3D wireframe extraction technique in multiple scenes. Fig.~\ref{fig:teaser} and Fig.~\ref{fig:applicationresults2} show two typical examples, a medium sized roughcast room and a large robotic lab. Both rooms were pre-scanned using the Faro 3D sensor. We simulate procedure 3D scanning by position a virtual camera at the center of the room and the camera rotates to capture a sequence of RGBD images. Fig.~\ref{fig:teaser} and Fig.~\ref{fig:applicationresults2} show that our system manages to extract accurate, complete and structural line segments in both cases, automatically forming high quality wireframe 3D models, a function largely missing in existing 3D scanning solutions. Recall that the Faro scanner still cannot recover regions occluded from the viewpoint of the scanning location, nor can it recover specular regions such as windows. The robotic lab scene is particularly challenging as direct mapping from 2D line segment to 3D introduces strong noise due to large depth range. Our technique, however, manages to not only reliably detect the structural and texture lines but also accurately determine their 3D locations. Fig.~\ref{fig:applicationresults2} (e) shows that our extracted 3D wireframe fits well with the point cloud. 

\section{Conclusions}
We have introduced a novel yet effective line segment detection method based on graph neural network. By representing each line segment as a quadruplet and all line segments in an image as a sparse graph, our method manages to not only extract structural line segments but also greatly reduce the computational cost. Benefiting from deep residual graph neural network, our method can effectively incorporate both semantic and geometric features of line segments.

Our future work will extend the LGNN in several directions. Firstly, with additional semantic annotations, we can jointly infer the geometric attribute and semantic label of each line segment for more coherent wireframe reconstruction. In addition, it is desirable to integrate line and plane detection for more robust 3D scene parsing. Furthermore, we will go beyond straight lines and consider other types of curved object boundaries in complex scenes.      

\begin{acks}
This work was supported by NSFC programs (61976138, 61977047), STCSM (2015F0203-000-06), the National Key Research and Development Program (2018YFB2100500) and SHMEC (2019-01-07-00-01-E00003).
\end{acks}

\bibliographystyle{ACM-Reference-Format}
\bibliography{reference}

\appendix

\end{document}